\pdfoutput=1
\PassOptionsToPackage{hyperfootnotes=false}{hyperref}
\documentclass[a4paper,fleqn]{cas-dc}

\usepackage[numbers,sort&compress]{natbib}
\usepackage{graphicx}
\graphicspath{{./}{figures/}}
\DeclareGraphicsExtensions{.pdf,.png,.jpg,.jpeg}
\usepackage{amsmath,amssymb}
\usepackage{booktabs}
\usepackage{multirow}
\usepackage{needspace}
\usepackage{placeins}
\usepackage{lastpage}
\usepackage{url}

\begin{document}
\def\lastpage{\pageref{LastPage}}
\RenewDocumentCommand{\printorcid}{}{}
\let\WriteBookmarks\relax
\def\floatpagepagefraction{1}
\def\textpagefraction{.001}

\shorttitle{Category Limits for Few-Shot Certification}
\shortauthors{Thai and Nguyen}

\title[mode = title]{How Many Categories Are Enough? Distribution-Free Certification Limits for Few-Shot Anomaly Thresholds}

\author[1]{Gia Huy Thai}
\ead{23120008@student.hcmus.edu.vn}

\author[2]{Nguyen Thai Anh}
\cormark[1]
\ead{anh.nt@vlu.edu.vn}

\affiliation[1]{organization={University of Science, VNU-HCM},
                city={Ho Chi Minh City},
                country={Vietnam}}

\affiliation[2]{organization={Faculty of Information Technology, Van Lang School of Technology, Van Lang University},
                city={Ho Chi Minh City},
                country={Vietnam}}

\cortext[cor1]{Corresponding author.}

\begin{abstract}
Few-shot anomaly detectors are judged by ranking metrics, yet deployment requires an alarm threshold with a controlled false-alarm rate (FAR). We ask how much normal evidence, in images or category units, is needed to certify such a threshold for an unseen category. Using a frozen DINOv2 principal component analysis (PCA) residual ranker on 15 MVTec and 12 VisA categories under four corruption types, we show that target-only leave-one-image-out (LOIO) calibration is resolution-limited and shift-fragile: rank values cannot fall below $1/(k+1)$, and at the attainable level $\alpha=0.20$, empirical FAR reaches 0.341 on Gaussian-corrupted MVTec at $k=4$, 1.7 times the nominal level. A category-count feasibility calculus is then derived: even with all-zero category losses and no multiplicity charged, any deterministic, uniformly valid, distribution-free 95\% upper confidence bound (UCB) requires at least 14, 29, and 59 independent and identically distributed (iid) category draws at $\alpha=0.20$, $0.10$, and $0.05$; these counts are necessary but not sufficient. The Cross-category Reliability Estimation with Source Support (CRESS) protocol splits source categories into disjoint reference, proposal, and certification roles. With only three or four certification categories, all 960 frozen configurations return the fail-closed threshold $\tau^\star=0$, and the smallest category-level UCB is 0.950. Image-unit analyses of the same archive select positive thresholds in 36.7\% to 60.3\% of target cells; these bounds hold for the selected source mixture, not for the marginal risk of a new-category draw. The contribution is a quantitative feasibility boundary and an estimand-aware protocol specifying when source evidence can, and cannot, support a transferable reliability claim.
\end{abstract}

\begin{keywords}
Industrial anomaly detection \sep Few-shot learning \sep Conformal prediction \sep Distribution-free certification \sep Distribution shift \sep Reliability
\end{keywords}

\maketitle
\hypersetup{
  pdftitle={How Many Categories Are Enough? Distribution-Free Certification Limits for Few-Shot Anomaly Thresholds},
  pdfauthor={Gia Huy Thai and Anh Nguyen},
  pdfsubject={Reliability and certification limits for few-shot industrial anomaly detection},
  pdfkeywords={industrial anomaly detection, few-shot learning, conformal prediction, distribution-free certification, distribution shift, reliability},
  pdfcreator={LaTeX using the Elsevier CAS double-column class}
}

\section{Introduction}\label{sec:introduction}
At the nominal false-alarm level $\alpha=0.20$, the target-only alarm evaluated in this study achieves an empirical false-alarm rate (FAR) of 0.341 on Gaussian-corrupted MVTec at $k=4$. This value represents the largest dataset-level aggregate in the frozen corruption grid and is approximately 1.7 times the intended level. This figure does not represent a population-level failure probability; rather, it is a benchmark observation pooled across categories and five support-sampling seeds under a single declared synthetic shift. However, this result highlights an operational discrepancy that ranking metrics alone do not capture. A detector may assign higher anomaly scores to defects than to normal images, resulting in a strong area under the receiver operating characteristic curve (AUROC) or average precision (AP), yet a fixed score threshold can still yield substantially different false-alarm behavior across categories or input conditions.

This discrepancy is especially significant in few-shot industrial anomaly detection (AD). Typically, a target category provides only $k\in\{1,2,4,8\}$ known-normal support images, while anomalous examples are rare, heterogeneous, and unavailable for threshold calibration. Frozen visual foundation models enable ranking in this context: DINOv2~\cite{oquab2024dinov2} patch features support nearest-neighbor (NN) memory banks, as implemented in PatchCore-style methods and AnomalyDINO~\cite{roth2022towards,damm2025anomalydino}, or compact principal component analysis (PCA) normal subspaces, as in SubspaceAD~\cite{subspacead2026}. However, these advances do not inherently determine which alarm threshold possesses a stable operational interpretation.

Two primary responses are possible. The first is to calibrate on the $k$ target-normal support images. The target-only method evaluates each support image using a leave-one-image-out (LOIO) fit and ranks a test score against the resulting normal scores. This approach is label-free and interpretable, but its rank values are restricted to the grid $\{j/(k+1)\}$. Increasing $k$ refines this grid; however, it does not guarantee that the support and shifted test scores meet the exchangeability conditions required for false-alarm validity. The empirical FAR of 0.341 illustrates this distinction: while the operating level is achievable, the shifted-normal rank distribution is anti-conservative compared to the exchangeable reference.

The second approach involves leveraging normal evidence from non-target source categories. A larger source archive provides finer empirical ranks and enables the proposal of thresholds below those available in the target-only grid. However, this strategy alters the underlying statistical question. Independent image samples from a specified source mixture can estimate risk for that mixture, whereas a guarantee for a previously unseen category requires independent and identically distributed (iid) category units from a defined meta-population. Repeated images, corruption views, or support seeds do not constitute new category units. Consequently, the central question addressed in this paper is:

\begin{quote}
How much source evidence, measured in which statistical unit, is needed to certify an anomaly threshold for a new category?
\end{quote}

We address this question at three levels. First, for any deterministic, uniformly valid, distribution-free 95\% upper confidence bound (UCB) based solely on independent and identically distributed (iid) category losses, an all-zero sample cannot certify $\alpha=0.20$, 0.10, or 0.05 with fewer than 14, 29, or 59 categories, respectively, even without multiplicity. These values represent optimistic impossibility limits rather than sufficient sample sizes for the procedure. Under frozen simultaneous allocation, the corresponding lower limits are at least 22, 46, and 94 for a single candidate, while the declared Hoeffding rule requires 60, 240, and 958 categories, respectively. Second, we audit target-only normal rank values under four controlled corruptions and demonstrate that crossing the finite resolution floor does not guarantee a stable FAR. Third, we employ CRESS to partition source categories into disjoint reference, threshold-proposal, and certification roles and assess whether the available archive achieves the predicted feasibility boundary.

CRESS functions as a source-assisted thresholding and audit protocol rather than as a novel anomaly ranker. The inherited ranking substrate is a frozen DINOv2 PCA residual. The reference categories establish a source-reference rank map, proposal categories define a finite candidate set, and certification categories evaluate fixed candidates using a family-adjusted UCB. The protocol outputs either the largest passing threshold or the fail-closed value of zero. Source-risk dominance enables the transfer of the bound to a specific target category. An iid category-sampling model provides only a marginal bound averaged over a new-category draw, rather than category-conditional control for each realized target. Neither normalization nor source pooling satisfies these requirements.

The strict audit includes only three certification categories in VisA cells and four in MVTec and transfer cells. This limitation is not solely attributable to the fixed $0.50/0.25/0.25$ split. The eligible source pools contain 11, 14, or 15 categories. Maintaining nonempty reference and proposal roles restricts certification categories to at most 9, 12, or 13, all below the optimistic 14-category threshold at $\alpha=0.20$. Consequently, category-unit certification assigns $\tau^\star=0$ to every target cell across the 960 frozen configurations. This outcome does not represent 960 independent failures or a computational or optimization failure. The configurations cover various jobs, source-view modes, support sizes, nominal levels, and conditions but share the same category-count obstruction. Their value is to show that the fail-closed outcome persists throughout the declared grid in the regime where analytic calculations predict infeasibility. In contrast, under the idealized iid source-image model, image-unit analyses include 72 to 191 units per cell and select positive thresholds in 36.7\% to 60.3\% of target cells, depending on the source-to-target job. This contrast does not reflect a simple stronger-versus-weaker certificate; rather, the two analyses address different risk profiles. The same source archive is statistically large when measured in images but insufficient when deployment is considered at the category level.

This paper presents three primary contributions:
\begin{itemize}
    \item \textbf{A category-count feasibility calculus.} The Hoeffding-specific requirement and a broader all-zero lower bound for deterministic uniformly valid distribution-free category-risk upper bounds are derived. These calculations quantify the minimum budget of iid category draws implied by the requested risk and confidence levels, serving as a study-design tool rather than solely a post hoc explanation.
    \item \textbf{An operational shift audit.} Across 15 MVTec and 12 VisA categories, the target-only analysis distinguishes unattainable nominal levels from attainable but unstable operating points. The largest dataset-level aggregate FAR is 0.341 on Gaussian-corrupted MVTec at $k=4$ and $\alpha=0.20$. VisA demonstrates conservative behavior in the corresponding cells, indicating that the shift effect is dataset-dependent rather than universal.
    \item \textbf{An estimand-aware source protocol.} The CRESS protocol separates reference, proposal, and certification categories, retains zero fallbacks, and reports image-mixture and category-level conclusions without substituting one for the other. The frozen strict audit verifies the predicted category shortage within datasets and for transfer from MVTec to VisA and the Metal Parts Defect Dataset (MPDD). Direct pooling and compact-ranker results are reported solely as supporting empirical context.
\end{itemize}

\begin{figure*}[pos=t]
    \centering
    \includegraphics[width=0.97\linewidth]{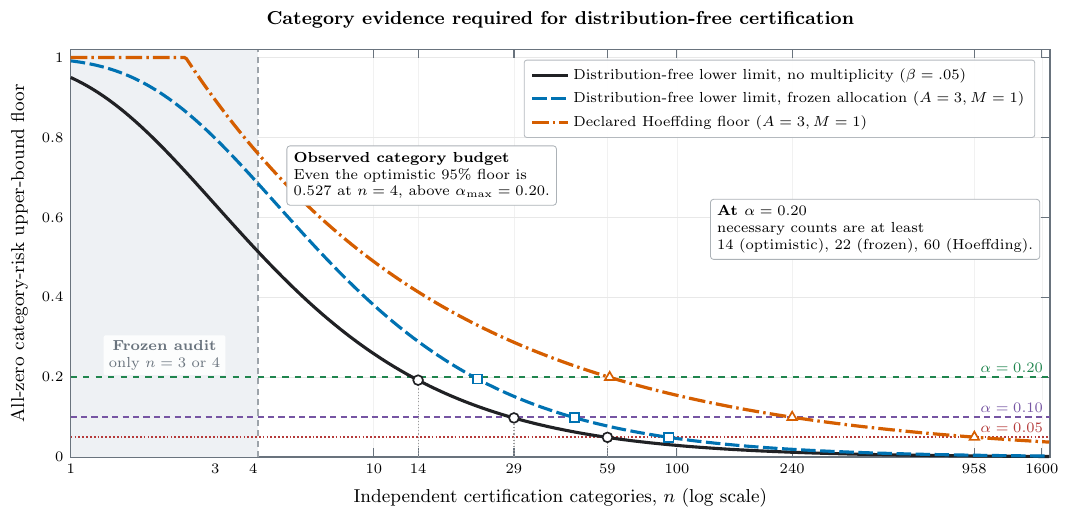}
    \caption{Analytic category-count feasibility after observing zero loss in every certification category. The black curve is the multiplicity-free lower bound $1-0.05^{1/n}$ for deterministic uniformly valid distribution-free procedures in Proposition~3. The blue curve applies the frozen allocation $\beta=\delta/(2AM)$ with $\delta=0.05$, $A=3$ reported levels, and the favorable one-candidate case $M=1$; the orange curve is the corresponding Hoeffding floor. Horizontal lines mark the requested risk levels, and the shaded region marks the three or four distinct certification categories available in the frozen audit; interpreting them as iid units is a modeling assumption. The black circular crossings at the 14, 29, and 59 axis ticks mark optimistic necessary counts without multiplicity, not sufficient sample sizes for CRESS. A curve above a requested $\alpha$ rules out a positive threshold only within its stated procedure class, even under all-zero observed category loss.}
    \label{fig:certificate-feasibility}
\end{figure*}

Figure~\ref{fig:certificate-feasibility} previews the central design calculation. The black curve represents the optimistic multiplicity-free floor for deterministic uniformly valid distribution-free bounds within the scope of Proposition~3. The blue and orange curves incorporate the frozen family allocation and the declared Hoeffding rule in their most favorable one-candidate case. A positive certificate is infeasible within the corresponding rule class whenever its curve remains above the requested $\alpha$. The shaded region indicates the three or four certification categories available in the strict audit. Table~\ref{tab:certificate-feasibility} subsequently reports the exact candidate-family calculations.

The remainder of the paper is organized as follows. Section~\ref{sec:related} positions the study against few-shot detectors, calibration, and conformal risk control. Section~\ref{sec:method} defines the inherited scorer, target-only route, source construction, CRESS protocol, and certificate scope. Section~\ref{sec:experiments} specifies the frozen datasets, statistical units, partitions, and claim rules. Section~\ref{sec:results} presents the shift audit, category-count feasibility, and strict CRESS, followed by supporting ranking and storage analysis. Sections~\ref{sec:limitations} and~\ref{sec:conclusion} outline the study's boundaries and implications.
\section{Related Work}\label{sec:related}
\subsection{Few-Shot Industrial AD}
Classical industrial AD typically assumed access to normal-only training data and identified deviations at test time. Patch distribution, self-supervised, and reconstruction-discriminative approaches such as PaDiM, CutPaste, and DRAEM provided important context for this literature~\cite{padim2021,cutpaste2021,draem2021}. PatchCore~\cite{roth2022towards} established memory-bank patch retrieval as a robust baseline by storing representative normal features and scoring test patches using NN distance. AnomalyDINO~\cite{damm2025anomalydino} extended this paradigm to few-shot inspection with frozen DINOv2 representations. Subsequent work introduced registration-based adaptation (RegAD~\cite{regad2022}), graph-structured memory (GraphCore~\cite{graphcore2023}), and fast feature reconstruction (FastRecon~\cite{fastrecon2023}). Vision-language methods included WinCLIP~\cite{winclip2023}, CLIP-SAM collaboration~\cite{clipsam2025}, and cross-modal prompt regularization~\cite{uniadprompt2026}. InCTRL introduced in-context residual learning~\cite{inctrl2024}. These methods were effective rankers; however, their raw score scales did not, by construction, provide a certified false-alarm rate interpretation. Calibration under distribution shift is also known to degrade for deep models in general~\cite{ovadia2019trust}. The present work is complementary: rather than competing on ranking, it examines how few-shot detectors should expose decision reliability.

\subsection{Frozen-Feature Subspace Detectors}
Subspace approaches replaced explicit patch retrieval with a compact normal subspace. SubspaceAD~\cite{subspacead2026} demonstrated that frozen DINOv2 patch features and PCA or subspace residuals were already effective for few-shot anomaly ranking, establishing the ranking mechanism as prior art. The present contribution begins after computing the residual and examines the reliability claims that a low-storage subspace detector can support under few-shot and distribution-shift conditions.

\subsection{Calibration and Decision Reliability in Few-Shot AD}
Calibration measured agreement between predicted confidence and empirical correctness~\cite{platt1999,guo2017calibration}. Standard post-hoc methods included Platt scaling and temperature scaling~\cite{platt1999,guo2017calibration}. Histogram binning and isotonic regression provided nonparametric alternatives~\cite{zadrozny2001obtaining,zadrozny2002transforming}. Sparse labels, open-ended anomaly types, and category-dependent scores made this objective especially difficult in AD. Prior reliability analyses of few-shot AD based on DINOv2 have also documented calibration and perturbation sensitivity~\cite{khan2025calibration}. These findings motivated our separation of ranking quality from the operational question of whether a declared alarm rule maintains its requested false-alarm behavior under a specified shift. A calibrated probability summary alone would not constitute such a risk guarantee.

\subsection{Conformal Prediction and Risk-Controlled Calibration}
Conformal prediction transformed nonconformity scores into p-values under exchangeability assumptions~\cite{vovk2005,shafer2008tutorial,angelopoulos2021gentle}. This approach was particularly well suited for normal-only few-shot AD because normal support images defined a reference distribution without needing real anomalies. The literature on conformal AD was well established. Leave-one-out, bootstrap, and cross-conformal anomaly detectors had been directly analyzed~\cite{hennhofer2024loio}. Weighted conformal detection in low-data regimes examined the effects of resolution and effective sample size~\cite{hennhofer2026weighted}. General nonconformity toolkits had been developed~\cite{hennhofer2026nonconform}. Few-shot conformal prediction with auxiliary tasks addressed the challenge of small target calibration sets~\cite{fisch2021fewshot}. Conformal AD, LOIO calibration, auxiliary-task transfer, and weighted conformal p-values were therefore established mechanisms.

Risk-controlling calibration methods were developed before this work. The Learn then Test framework formulated finite-sample calibration as a multiple-testing problem over candidate configurations~\cite{angelopoulos2021learntest}. Conformal risk control selected parameters to control the expectation of a monotone loss~\cite{angelopoulos2024crc}. In anomaly detection, PAC-Wrap provided semi-supervised probably approximately correct (PAC) bounds for false-positive and false-negative rates around a base detector~\cite{li2022pacwrap}. These studies established general principles of candidate testing, held-out risk assessment, and one-sided error bounds. This work does not claim those statistical primitives as novel. Instead, CRESS applies these principles for normal-only cross-category thresholding by assigning disjoint source categories to reference, proposal, and certification roles. It distinguishes selected-mixture image risk from marginal new-category risk and audits whether the available category count can support a nonzero threshold under the declared iid model.

Structure-aware conformal adaptation has also been studied outside the context of anomaly detection. Lam and Anh proposed Heterophily-Aware Diffused Conformal Prediction (HeAD-CP), which adapted graph-score diffusion using label-free local homophily estimates and showed that uniform diffusion could impair prediction-set efficiency on heterophilic graphs while preserving marginal coverage~\cite{lam2026headcp}. This result supported the broader need to audit reliability transformations under structural shift; however, its setting concerned graph neural network (GNN) node-classification sets rather than few-shot industrial anomaly alarms or cross-category source certification.

The deterministic source-view choices were conceptually informed by the shared-versus-specialized selection motif in Shape-Adapting Gated Experts (SAGE)~\cite{Thai_2026_CVPR}. That work concerned learned expert selection rather than certification of anomaly thresholds. CRESS does not use a learned expert gate; its source-view modes are predeclared protocol choices, and its reference, proposal, and certification partitions have disjoint statistical roles. The citation therefore records design inspiration without attributing CRESS validity or novelty to the SAGE architecture.

This contribution begins with the independent-category requirement of a simultaneous source certificate, using the target-only grid $1/(k{+}1)$ as a motivating finite-resolution constraint. Established confidence tools are specialized into Hoeffding-specific and distribution-free all-zero category-budget calculations. Corruption-shift deviations are empirically audited. Source-assisted candidate thresholds are evaluated without target anomaly labels. CRESS does not introduce a new nonconformity score or generic confidence bound. Instead, it specifies the separation of source categories into reference, proposal, and certification roles and audits whether the resulting source certificate can be nonvacuous at the available category count.

\section{Method}\label{sec:method}
\subsection{Problem Setup}
For each target category $c$, a support set $\mathcal{S}_c=\{x_i\}_{i=1}^{k}$ consisting solely of normal images is observed. Target test images may be either normal or anomalous, but their labels are reserved exclusively for evaluation and are not utilized in detector fitting, threshold proposal, or certification. Other categories may additionally provide a declared source-normal archive. The framework outputs a fixed raw anomaly score $s_c(x)$ and an optional pixel map, a target-only reliability value $p_{\mathrm{LOIO}}(x)$ when LOIO calibration is available, and, for the source-assisted route, a threshold $\tau^\star$ with an accompanying source-domain certificate. Figures~\ref{fig:target-pipeline} and~\ref{fig:cress-pipeline} distinguish the inherited ranking substrate, the target-only route, and the source-assisted route.

The statistical unit is fixed before threshold evaluation. Under the idealized iid image model, image-unit analysis gives equal weight to each retained source image and targets the mean alarm loss for a sample drawn from the induced source-image mixture, where category weights are proportional to the number of retained images. In contrast, category-unit analysis first averages alarms within each source category, gives equal weight to each certification category, and then targets the expected archive-level loss for a new sample from the declared source-category meta-population. In this context, a category unit consists of the category, its support-set construction, and its predeclared held-out normal-view sampling mechanism. To equate this expectation with the alarm probability of a fresh normal image, the within-category views must be representative of the declared deployment distribution. Independent category units remain necessary; additional images, support seeds, or corruption views within an existing category do not increase the number of units. Neither source estimand directly corresponds to the target-category risk. Achieving this transfer requires the additional condition specified at the end of this section.

\subsection{Frozen Subspace Scoring and Localization}
Patch features are extracted using a frozen DINOv2 small Vision Transformer with $14\times14$ patches (ViT-S/14). Each RGB image is resized to $518\times518$ pixels and normalized using the ImageNet channel mean and standard deviation. The normalized patch-token output contains a $37\times37$ grid of $d=384$ vectors. Let $z_{ij}\in\mathbb{R}^{d}$ represent patch $j$ from image $i$. A PCA subspace is fitted on all target-support patches. Given the support mean $\mu_c$ and retained principal directions $U_c\in\mathbb{R}^{d\times m}$, the patch residual is defined as
\begin{equation}\label{eq:residual}
    r_c(z)=\left\|(z-\mu_c)-U_cU_c^{\top}(z-\mu_c)\right\|_2.
\end{equation}
The patch residuals are used to construct an anomaly map. For the conformal protocols, the $N=1369$ residuals are arranged in descending order as $r_{c,(1)}(x)\geq\cdots\geq r_{c,(N)}(x)$, and the image-level ranking score is defined as follows:
\begin{equation}\label{eq:score}
    s_c(x)=\frac{1}{q}\sum_{j=1}^{q}r_{c,(j)}(x),
    \qquad q=\lceil0.01N\rceil=14.
\end{equation}
This top-$1\%$ mean serves as a smoothed maximum. In contrast, the clean accuracy-storage benchmark uses the patch maximum ($q=1$). AUROC and AP are always computed from the fixed score $s_c(x)$ within a specified protocol; subsequent reliability transformations do not alter this score or its ranking. Certification in this study pertains to image-level alarms rather than pixel-map thresholds.

\subsection{Target-Only LOIO Reliability}
For $k \geq 2$, the target-only approach constructs normal calibration residuals using LOIO support splitting. For each support image $x_i$, a subspace is fitted on $\mathcal{S}_c \setminus \{x_i\}$, and the held-out image is assigned the score $s_{c,-i}(x_i)$. The resulting set
\begin{equation}\label{eq:rloio}
    \mathcal{R}_{c,\mathrm{LOIO}}
    =\{s_{c,-i}(x_i):x_i\in\mathcal{S}_c\}
\end{equation}
serves as target-normal nonconformity evidence. A test image, scored using the subspace fit on all $k$ supports, receives the conformal-form rank value
\begin{equation}\label{eq:loio}
    p_{c,\mathrm{LOIO}}(x)
    =\frac{1+\sum_{r\in\mathcal{R}_{c,\mathrm{LOIO}}}
    \mathbf{1}\{r\geq s_c(x)\}}
    {1+|\mathcal{R}_{c,\mathrm{LOIO}}|}.
\end{equation}

The conventional $p$ notation is retained but Equation~\eqref{eq:loio} is referred to as a rank value unless its validity conditions are explicitly invoked. This metric quantifies extremeness relative to held-out support normals. Neither this value nor $1-p_{c,\mathrm{LOIO}}(x)$ represents a supervised anomaly posterior.

This construction is asymmetric: each calibration score utilizes $k-1$ support images, while the test score employs all $k$. Consequently, Equation~\eqref{eq:loio} serves as a LOIO approximation, consistent with the principles of cross-conformal inference~\cite{vovk2005,barber2021jackknife,hennhofer2024loio}, rather than a split-conformal p-value that provides an automatic finite-sample guarantee~\cite{bates2023testing,laxhammar2015inductive}. A matched-LOIO audit evaluates held-out support and test images using the same fold-specific subspaces and is presented as a sensitivity analysis. The construction is maintained as declared, and its normal rank values are empirically audited rather than relying on the assumption of exchangeability.

\begin{figure*}[t]
\centering
\includegraphics[width=0.98\textwidth]{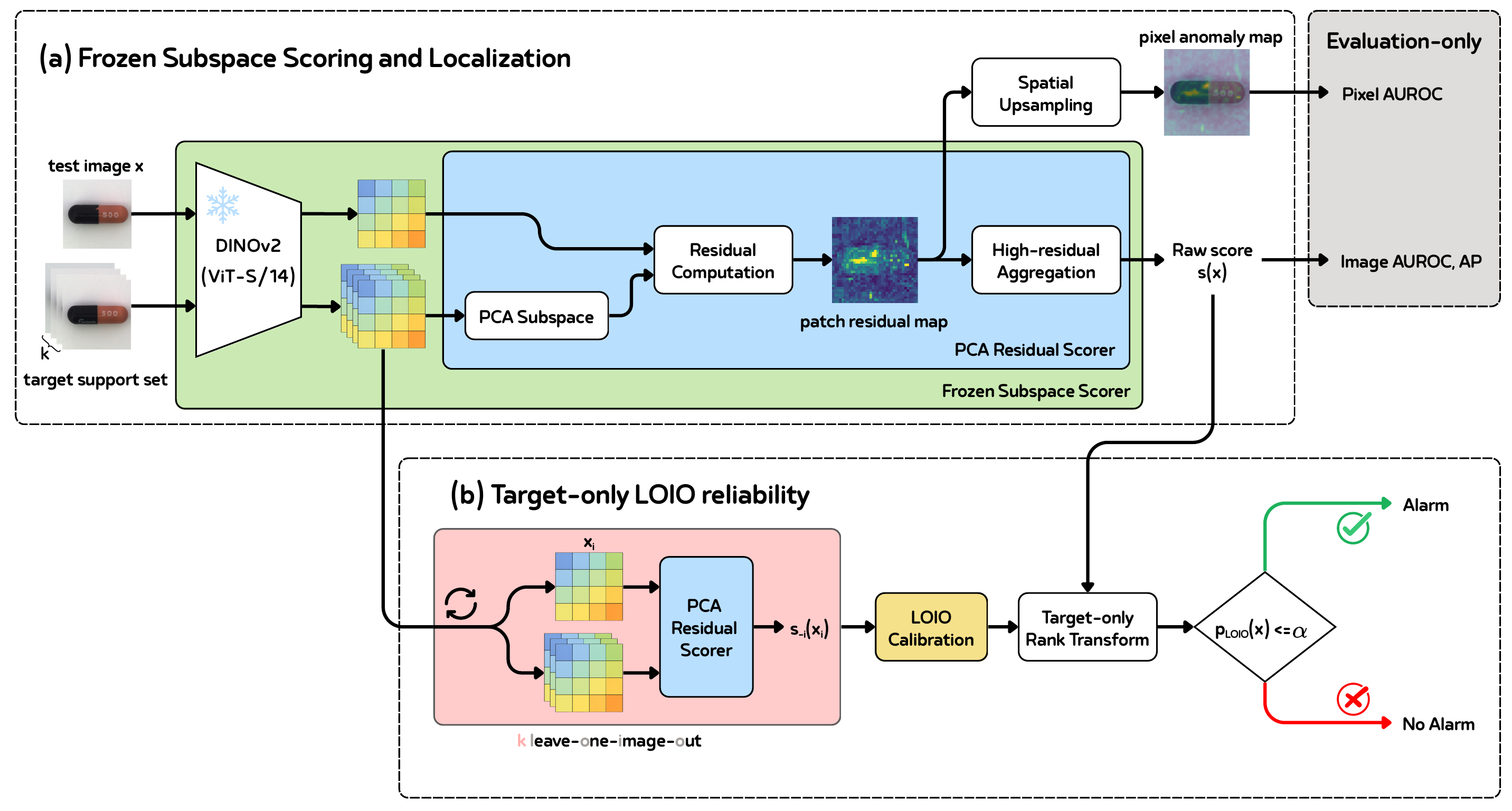}
\caption{Target-side scoring and reliability routes. Panel (a) employs frozen DINOv2 patch features and a support-fitted PCA residual scorer to produce the fixed image-ranking score and, optionally, a pixel anomaly map. The metric boxes are used exclusively for evaluation; the pixel map is an inherited output and does not serve as a certification target or a localization claim within this study. Panel (b) applies the scorer within LOIO folds to generate target-normal calibration scores and the target-only rank value $p_{c,\mathrm{LOIO}}(x)$. Since calibration scores utilize $k-1$ supports and the test score uses all $k$, this rank transformation is subject to empirical auditing rather than being assumed to possess automatic finite-sample validity. The minimum attainable value is $1/(k+1)$, and the alarm rule does not modify the underlying ranking score.}
\label{fig:target-pipeline}
\end{figure*}

Figure~\ref{fig:target-pipeline} provides an overview of the target-side workflow. Panel (a) illustrates the shared frozen subspace scoring and localization substrate, while panel (b) demonstrates the reuse of the same scorer to construct target-only LOIO rank values.

\subsection{Target-Only Resolution}
\noindent\textbf{Remark 1 (attainable-alpha floor).} For any finite calibration scores $r_1,\ldots,r_k$, a rank value of the form in Equation~\eqref{eq:loio} belongs to $\{j/(k{+}1):j=1,\ldots,k{+}1\}$. Hence the alarm $\mathbf{1}\{p(x)\leq\alpha\}$ is identically zero for $\alpha<1/(k{+}1)$.

\noindent\emph{Proof.} The exceedance count is an integer from 0 to $k$. Adding one and dividing by $k+1$ gives the stated grid and lower bound. No exchangeability assumption is needed for this algebraic claim. \hfill$\square$

For $k=4$, the floor is $0.2$; for $k=8$, it is $1/9\approx0.111$. The attainable-alpha analysis reports, for each nominal $\alpha$, the largest attainable grid point not exceeding $\alpha$ and the corresponding empirical FAR. The floor represents a standard finite-rank resolution property~\cite{hennhofer2026weighted}, rather than an algorithmic innovation or evidence that the underlying ranker lacks the ability to separate anomalies. While randomization can smooth this deterministic grid under the relevant exchangeability assumptions, it cannot, by itself, address support-to-test shift.

\subsection{Source Archive Preparation}\label{sec:cress-source-construction}
CRESS utilizes known-normal images from categories distinct from the target. For each target or source category $c$, $\mathcal{B}_c$ denotes its support-only calibration residuals. For $k\geq2$, $\mathcal{B}_c=\mathcal{R}_{c,\mathrm{LOIO}}$. Since LOIO is undefined with a single support image, the specified $k=1$ stress test employs one patch-split support calibration residual. The set $\mathcal B_c$ is summarized by its median and median absolute deviation (MAD):
\begin{equation}\label{eq:cress-normalization}
\begin{aligned}
    m_c &= \operatorname{median}(\mathcal{B}_c),\\
    d_c &= \operatorname{MAD}(\mathcal{B}_c),\\
    \widetilde{s}_c(x) &= \frac{s_c(x)-m_c}{\max(d_c,\varepsilon)}.
\end{aligned}
\end{equation}
Here $\varepsilon=10^{-6}$ is fixed prior to evaluation. This support-only normalization improves comparability of score scales across categories, but does not imply equality or exchangeability of their distributions. Specifically, when $k=1$, the single calibration residual yields zero MAD, so the denominator defaults to the $\varepsilon$ threshold. These cases are interpreted as stress tests rather than standard LOIO calibration.

Normalized source score views are selected without reference to target labels or target scores. In matched-condition mode, the source view corresponding to the declared target condition is selected, which requires condition metadata. Clean-source mode consistently selects clean source normals. In condition-agnostic mode, for each base source image, the median normalized score across all available condition views is used to prevent repeated corruptions from being treated as independent observations. Deterministic mismatched-condition routing employs the lexicographic successor of the target condition, with wraparound, as a negative control. Normalization is performed prior to source-view selection or aggregation in the protocol. Only the resulting scalar score archive and category identities are required for downstream processes; source patch features are excluded from certification.

For a fixed target and support seed, eligible source categories are split before candidate construction into disjoint reference, proposal, and certification sets $\mathcal{R}$, $\mathcal{P}$, and $\mathcal{C}$. The primary allocation ratio is $0.50/0.25/0.25$; integer counts are assigned using the largest fractional remainders, with ties resolved in favor of certification, ensuring that each partition is nonempty. The category permutation is determined by a hash of the target-category identity and seed, remains constant across conditions and $k$, and is independent of observed scores, target outcomes, or certification losses. All score views from a given category are assigned to the same partition.

\subsection{Nested CRESS Threshold Selection}\label{sec:cress-threshold-selection}
Let $\mathcal{Z}_{\mathcal{R}}$ be the pooled normalized scores from the selected reference-category views. The reference categories define the source-reference rank map
\begin{equation}\label{eq:cress-source-pvalue}
    p_{\mathcal{R}}(x)
    =\frac{1+\sum_{r\in\mathcal{Z}_{\mathcal{R}}}
    \mathbf{1}\{r\geq\widetilde{s}_c(x)\}}
    {1+|\mathcal{Z}_{\mathcal{R}}|}.
\end{equation}
The term \emph{source-reference map} is used deliberately: Equation~\eqref{eq:cress-source-pvalue} represents an empirical rank transformation on the source archive, and cross-category conformal validity is not assumed. The finer grid results from the larger reference-score pool rather than from the $k$ target supports.

Applying the reference map to the proposal categories yields the proposal rank values ${p_{\mathcal R}(x):x\in\mathcal P}$. The distinct values of these ranks generate an ordered candidate set $\mathcal{T}$. The realized size is $M=|\mathcal T|$; if this exceeds $M_{\max}=20$, a deterministic, evenly spaced subset is retained. Certification data do not influence either $\mathcal T$ or $M$. The same reference map is subsequently applied to every certification image. For $\tau\in\mathcal T$, the image-unit loss is defined as
\begin{equation}\label{eq:cress-image-loss}
    L_x(\tau)=\mathbf{1}\{p_{\mathcal{R}}(x)\leq\tau\}.
\end{equation}
All certification images are declared normal; therefore, any alarm constitutes a false alarm. For a certification category $c$, the category-unit loss is
\begin{equation}\label{eq:cress-category-loss}
    L_c(\tau)=\frac{1}{n_c}\sum_{x\in c}L_x(\tau)\in[0,1].
\end{equation}
Under iid image sampling, the image unit targets the mean loss for the declared source-image mixture. In contrast, the category unit targets the expected archive-average loss for a complete category unit drawn iid from the declared meta-population. Treating $L_c(\tau)$ as a single bounded unit permits arbitrary dependence among its archived views; however, Hoeffding's inequality requires iid sampling across complete category units. If the within-category archive accurately represents the declared deployment distribution, the category expectation also equals the marginal alarm probability after drawing a category, its support, and a new normal view. Otherwise, it remains an archive-level estimand. The image and category estimands are not interchangeable.

Let $A$ be the number of reported nominal levels, let $\delta\in(0,1)$ be the per-cell family failure-probability budget, and let $n=|\mathcal{C}|$ be the number of certification categories. For each candidate threshold, define the empirical mean category loss
\begin{equation}\label{eq:cress-mean-category-loss}
    \overline{L}_{\mathcal C}(\tau)=\frac{1}{n}\sum_{c\in\mathcal{C}}L_c(\tau).
\end{equation}
The category analysis then uses the family-adjusted Hoeffding category-risk UCB
\begin{equation}\label{eq:cress-ucb}
    U_{\mathrm{H}}(\tau)=\min\!\left\{1,\;
    \overline{L}_{\mathcal C}(\tau)+
    \sqrt{\frac{\log(2AM/\delta)}{2n}}\right\}.
\end{equation}

For image-unit sensitivity, pooled alarm indicators are modeled as iid Bernoulli draws from the induced source-image mixture, conditional on the fitted source scorers. The stratified archive does not guarantee this idealized model. The exact one-sided Clopper-Pearson upper bound is used with tail probability $\delta/(2AM)$. The factor $2A$ distributes the family error across both unit definitions and reported levels, while the realized $M$ allocates it across candidates within that cell. In the selection rule below, $U$ denotes $U_{\mathrm H}$ for category certification and the corresponding Clopper-Pearson bound for image-unit sensitivity. CRESS returns
\begin{equation}\label{eq:cress-selected-threshold}
    \tau^\star=
    \max\bigl(\{\tau\in\mathcal{T}:U(\tau)\leq\alpha\}\cup\{0\}\bigr).
\end{equation}
The value $\tau^\star=0$ serves as a deterministic fail-closed fallback since $p_{\mathcal{R}}(x)>0$. After threshold selection, a target image is processed by
\begin{equation}\label{eq:cress-target-decision}
    a_{\tau^\star}(x)
    =\mathbf{1}\{p_{\mathcal{R}}(x)\leq\tau^\star\},
\end{equation}
where the target score is normalized only by the target support statistics. Target test images and labels enter none of the reference, proposal, certification, or threshold-selection stages.

\begin{figure*}[t]
\centering
\includegraphics[width=0.98\textwidth]{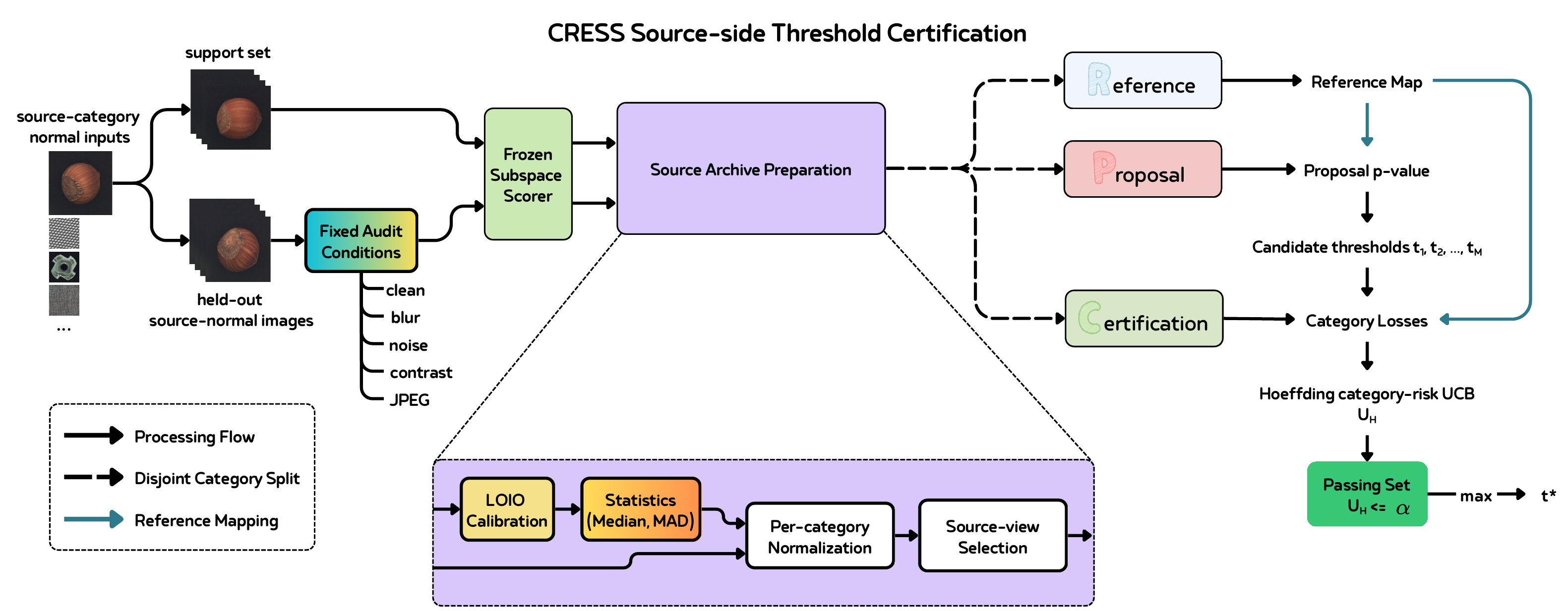}
\caption{CRESS source-side threshold certification. Each non-target source category provides normal support images for scorer fitting and support-only normalization statistics, along with held-out known-normal views under specified audit conditions. During source-archive preparation, scores are normalized before source-view selection. When $k=1$, the declared patch-split support residual replaces LOIO calibration. Eligible source categories are partitioned into disjoint reference ($\mathcal R$), proposal ($\mathcal P$), and certification ($\mathcal C$) roles. The reference set establishes the source-reference map $p_{\mathcal R}$, which is then applied to proposal and certification views. The proposal set determines the candidate family, while the certification set generates category losses for the family-adjusted risk UCB. CRESS returns either the largest passing threshold or the fail-closed value zero. This process yields a source-domain certificate; however, target-category control requires the transfer condition specified in Corollary~1.}
\label{fig:cress-pipeline}
\end{figure*}

Figure~\ref{fig:cress-pipeline} provides an overview of the complete CRESS source-side workflow, including source-archive preparation, disjoint category splitting, candidate proposal, category-risk certification, and fail-closed threshold selection.

\subsection{Certificate Scope, Feasibility, and Transfer Boundary}
\noindent\textbf{Proposition 1 (Conditional Source Certificate).} Consider a fixed target, support size, seed, source-view mode, and condition, with conditioning on both the reference and proposal stages. Under the image model, further condition on the fitted certification scorers. Assume that certification units are independent samples from the declared source-unit population: either pooled Bernoulli alarm indicators in the image model, or category units that include their support construction and held-out views and yield bounded losses in $[0,1]$ in the category model. Then, with probability at least $1-\delta$, simultaneously across all $A$ levels and both unit definitions, the source risk corresponding to every selected threshold does not exceed its $\alpha$.

\noindent\emph{Proof.} For a fixed level, unit, and candidate, the one-sided Clopper–Pearson construction or Hoeffding's inequality fails with probability at most $\delta/(2AM)$. Applying a union bound over $M$ candidates, $A$ levels, and two unit types yields a total failure probability of at most $\delta$. Independence between these bounds is not required. Selection is performed only on this simultaneous event, and the zero fallback results in zero alarm risk. \hfill$\square$

\noindent\textbf{Proposition 2 (Hoeffding Feasibility).} Under the category-level Hoeffding rule, a positive candidate passes at level $\alpha$ only if
\begin{equation}\label{eq:cress-feasibility}
    n\geq\frac{\log(2AM/\delta)}{2\alpha^2}.
\end{equation}
\noindent\emph{Proof.} Since $\overline{L}_{\mathcal C}(\tau)\geq0$, the unclipped upper bound is at least $\sqrt{\log(2AM/\delta)/(2n)}$. Imposing the requirement that this bound does not exceed $\alpha$ and rearranging the inequality establishes the claim. \hfill$\square$

\noindent\textbf{Proposition 3 (distribution-free all-zero lower bound).}
Let $\beta\in(0,1)$ and let $L_1,\ldots,L_n$ be iid losses from an arbitrary distribution $P$ on $[0,1]$, with mean $\mu_P$. Let $U_n:[0,1]^n\to[0,1]$ denote any deterministic UCB satisfying
\[
    \inf_{P}\Pr_{P^n}\!\left\{\mu_P\leq U_n(L_1,\ldots,L_n)\right\}\geq1-\beta.
\]
Then,
\begin{equation}\label{eq:distribution-free-feasibility}
    U_n(0,\ldots,0)\geq1-\beta^{1/n}.
\end{equation}
Consequently, even if zero loss is observed in every certification category, $U_n\leq\alpha$ is possible only if
\[
    n\geq\frac{\log\beta}{\log(1-\alpha)}.
\]

\noindent\emph{Proof.}
Let $u=U_n(0,\ldots,0)$ and assume, for contradiction, that $u<1-\beta^{1/n}$. Select a Bernoulli distribution with mean $q$ such that $u<q<1-\beta^{1/n}$. The probability of observing an all-zero sample is $(1-q)^n>\beta$. On this event, $U_n=u<q=\mu_P$, so the bound fails with probability greater than $\beta$, which contradicts the requirement of uniform $(1-\beta)$ coverage. Rearranging $1-\beta^{1/n}\leq\alpha$ yields the sample-size condition. \hfill$\square$

Proposition~3 demonstrates that the frozen failure is not merely a consequence of Hoeffding looseness. The Bernoulli counterexample is straightforward; the primary contribution is the application of this necessary condition as an explicit category-budget calculation for transferable anomaly thresholds. Even with a single fixed candidate, one level, no image-unit co-reporting, and no multiplicity penalty ($\beta=\delta=0.05$), any deterministic, uniformly valid, distribution-free bound based only on iid category losses requires at least 14, 29, and 59 categories at $\alpha=0.20$, $0.10$, and $0.05$, respectively. These values represent necessary lower bounds, rather than assertions that the corresponding sample sizes are sufficient for CRESS. Under the frozen allocation $\beta=\delta/(2AM)$ with $A=3$ and the most favorable $M=1$, the corresponding lower bounds are 22, 46, and 94; the stated Hoeffding rule requires 60, 240, and 958.

An equivalent design interpretation is obtained by solving Equation~\eqref{eq:distribution-free-feasibility} for the confidence level. With $n$ all-zero categories, no bound within the scope of Proposition~3 can certify risk at most $\alpha$ with confidence exceeding $1-(1-\alpha)^n$. For $\alpha=0.20$, three and four categories support at most 48.8\% and 59.0\% confidence, respectively, in the optimistic multiplicity-free scenario. Conversely, at 95\% confidence, their all-zero UCB floors are 0.632 and 0.527. This formulation quantifies the attainable confidence given the available category budget, without implying that this budget suffices for CRESS.

\begin{table}[t]
\centering
\caption{The minimum iid certification-category counts required by the all-zero calculations at $\delta=0.05$. The distribution-free rows indicate the necessary lower bounds for deterministic, uniformly valid procedures as described in Proposition~3, first without multiplicity ($\beta=\delta$), and then with $\beta=\delta/(2AM)$, $A=3$, and realized family size $M$. The 'Hoeffding' row corresponds to Proposition~2. All rows assume zero observed loss; satisfying a listed count does not guarantee a positive CRESS threshold.}
\label{tab:certificate-feasibility}
\footnotesize
\setlength{\tabcolsep}{4pt}
\begin{tabular}{lrrr}
\toprule
rule / candidates & $\alpha=.20$ & $\alpha=.10$ & $\alpha=.05$ \\
\midrule
Distribution-free, no multiplicity & 14 & 29 & 59 \\
\midrule
Distribution-free, $M=1$  & 22 & 46 & 94 \\
Distribution-free, $M=5$  & 29 & 61 & 125 \\
Distribution-free, $M=20$ & 35 & 74 & 152 \\
\midrule
Hoeffding, $M=1$  & 60 & 240 & 958 \\
Hoeffding, $M=5$  & 80 & 320 & 1,280 \\
Hoeffding, $M=20$ & 98 & 390 & 1,557 \\
\bottomrule
\end{tabular}
\end{table}

Table~\ref{tab:certificate-feasibility} presents a design calculation rather than an empirical result. Proposition~3 applies only to deterministic, uniformly valid, distribution-free upper bounds based exclusively on iid bounded category losses. Parametric or hierarchical assumptions, informative auxiliary data, and randomized confidence procedures necessitate separate validity justifications. While adding independent source-image draws may enhance the selected-mixture image analysis, additional images or repeated support seeds do not increase the number of independent certification categories. Similarly, alternative disjoint allocations cannot recover the current source pools while maintaining nonempty $\mathcal R$ and $\mathcal P$: pools of 14, 11, and 15 eligible categories allow at most 12, 9, and 13 certification categories, respectively, each below the optimistic multiplicity-free requirement of 14 at $\alpha=0.20$.

\noindent\textbf{Corollary 1 (conditional target transfer).} Under the event described in Proposition~1, the risk of a specific target category under the evaluated support construction does not exceed $\alpha$ if, for every proposed threshold, its normal alarm probability is no greater than the corresponding certified source risk. Alternatively, for the category-unit bound, if a target category, along with its support construction and normal-view sampling mechanism, constitutes an independent draw from the same meta-population as the certification-category units, then the archive-level alarm loss, marginalized over that random category draw, is at most $\alpha$. Under representative within-category sampling, this statement also applies to the marginal alarm probability of a new normal view.

\noindent\emph{Proof.} The first statement is a consequence of source-risk dominance and Proposition~1. If the complete target unit is an independent draw from the same meta-population, its marginal archive-level risk matches the source meta-population risk bounded in Proposition~1. Representative within-category sampling equates the expectation of the archive average with the corresponding fresh-view alarm probability. Neither argument provides a bound on the category-conditional risk for every realized target. \hfill$\square$

Source-risk dominance and iid category sampling are assumptions rather than consequences of normalization, source-view selection, or the rank transformation in Equation~\eqref{eq:cress-source-pvalue}. Therefore, $\tau^\star$ provides a source-domain certificate. It serves as a guarantee for a specific target only under the dominance assumption; iid category sampling yields only the weaker marginal new-category statement. The direct pooled-source conformal approach, which utilizes the complete selected source pool with $\tau=\alpha$ and omits a disjoint proposal or certification stage, is maintained as an uncertified empirical reference.

\section{Experiments}\label{sec:experiments}
\subsection{Datasets and Few-Shot Splits}
MVTec AD~\cite{mvtec2019} and VisA~\cite{visa2022} are evaluated in this study. The strict external-transfer audit also incorporates MPDD. MVTec AD includes industrial object and texture categories with pixel-level defect masks. VisA encompasses a broader range of industrial categories and serves as the primary benchmark for full-scale corruption and calibration shift. MPDD provides six metal-part categories for an external stress test that transfers from MVTec to MPDD. Within-MPDD certification is not conducted because the frozen nested implementation requires at least six eligible source categories, and excluding one MPDD target leaves only five. For each category and seed, a random permutation of the path-sorted normal training images defines nested support sets, ensuring that the $k$-shot set is contained within every larger declared set. The CRESS source archive is maintained separately and consists of held-out known-normal evaluation images from non-target source categories (the \texttt{test/good} partition for MVTec and the corresponding normal partition for VisA and MPDD). Source normal status is used, but target test labels are never included in reference construction, threshold proposal, or certification. Unless otherwise specified, $k\in\{1,2,4,8\}$ is used for the clean accuracy-storage and strict nested studies, while the target-only corruption tables use $k\in\{4,8\}$. The $k=1$ strict cells employ the declared patch-split support calibration fallback and are interpreted as stress tests; LOIO calibration requires $k\geq2$.

\subsection{Evaluation Protocols}
The target-only corruption audit evaluates all test images from 15 MVTec and 12 VisA categories at $k\in\{4,8\}$ under five support-sampling seeds, with the $k$ sets nested within each seed and four corruption types. Support images remain clean, and each corruption is applied only to evaluation views. Images are represented on $[0,1]$. Gaussian noise introduces independent $\mathcal N(0,0.05^2)$ pixel-channel noise and clips the result. Blur is implemented using a box-filter radius of one pixel, corresponding to the declared three-pixel kernel. Brightness and contrast adjustments apply $\operatorname{clip}(1.15(x-0.5)+0.55,0,1)$. JPEG corruption uses a quality-60 encode/decode process. Each transformed array is clipped, quantized to 8-bit, and stored as a PNG before feature extraction. The noise generator uses the support seed plus the fixed image index, while the other transformations are deterministic. Primary rank values follow Equation~\eqref{eq:loio}: calibration scores use LOIO fits, and each test score uses the full-support fit. The fold-matched construction is reported only as a sensitivity analysis. Alarm comparisons use the predeclared tolerance $10^{-6}$ to absorb fp32 representation error at attainable grid points.

The primary empirical groups include the target-only corruption audit and the frozen strict nested CRESS audit, while the category-count feasibility curves are based on analytic design calculations. Supporting groups address clean ranking and storage, PCA64/PCA128 sensitivity, direct pooled-source conformal, and attainable operating levels. The strict nested protocol employs $k\in\{1,2,4,8\}$, seeds from 0 to 4, $\alpha\in\{0.05,0.10,0.20\}$, and clean, Gaussian-noise, blur, brightness/contrast, and JPEG conditions. The shared score export retains at most 120 evaluation records per category, seed, and condition through deterministic label-stratified sampling, with dataset labels determining the retained evaluation strata. Target labels are used only to score evaluation outcomes and do not enter threshold construction. Downstream source construction discards every anomalous source row, so source anomaly scores are excluded from the reference map, proposal, and certification losses. The remaining cells use $\rho=0.01$, PCA64, $\delta=0.05$, and realized $M=|\mathcal T|\leq20$. Raw scores are first normalized using support-only statistics, and condition-specific source views are then selected or aggregated as defined in Section~\ref{sec:cress-source-construction}. For each target category and seed, eligible source categories are deterministically split using the frozen $0.50/0.25/0.25$ reference/proposal/certification allocation. This results in 7/3/4 categories for a 14-category within-MVTec source pool, 5/3/3 for an 11-category within-VisA pool, and 7/4/4 for the 15-category MVTec transfer pool. The same category partition is reused across conditions and $k$. The protocol evaluates MVTec and VisA within-dataset, MVTec-to-VisA, and MVTec-to-MPDD under matched-condition, clean-source, condition-agnostic, and mismatched negative-control source selection.

The frozen operational gate requires a positive category-level threshold in at least 80\% of target cells. It subsequently checks target FAR, power, no-harm relative to LOIO, and power gain below the target-only floor. The 80\% cutoff serves as a predeclared usability criterion rather than a confidence level; since the observed rate is zero, any positive cutoff yields the same verdict. The confidence allocation described in Proposition~1 applies within each fixed target, seed, source-view mode, and condition cell, but does not constitute a simultaneous 95\% guarantee over the entire experimental grid. Failed cells are retained.

\subsection{Baselines and Ablations}
The inherited PCA residual scorer is compared with a controlled nearest-neighbor memory bank, both using the same frozen DINOv2 features and few-shot split. PCA64 and PCA128 retain 64 and 128 principal directions, respectively, with PCA64 serving as the default reliability substrate. In the source-assisted study, direct pooled-source conformal serves as an uncertified baseline; it employs the complete selected source pool with $\tau=\alpha$, but does not perform reference, proposal, or certification splits and does not apply a UCB gate. The controlled memory-bank implementation is not an official reproduction of PatchCore or AnomalyDINO, but instead isolates scoring and storage trade-offs under a unified protocol. AnomalyDINO-S (448)~\cite{damm2025anomalydino} is included based on values reported under its published protocol. WinCLIP~\cite{winclip2023} is also included only through published values, as the audit identifies no author-released implementation associated with the cited paper. SubspaceAD is treated as a prior-art guardrail rather than a controlled baseline; the frozen DINOv2 subspace residual is inherited, and the contribution begins with the reliability analysis downstream of that score.

\subsection{Metrics}
Ranking metrics are distinguished from reliability metrics. Image AUROC and AP are calculated from the raw PCA residual score $s(x)$; pixel maps are not considered certification targets. Reliability is evaluated operationally by empirical FAR at fixed $\alpha$, detection power, alarm precision, the empirical cumulative distribution function (CDF) of normal rank values, and the fraction of cells receiving a positive certified threshold. Storage is quantified as retained ranker state per category. Corruption views serve as controlled stress tests and are not treated as independent images or as proxies for all deployment shifts.

\paragraph{Category-level residual-displacement analysis.}
To examine the difference between MVTec and VisA without treating images or support seeds as independent replications, a paired category-level analysis is conducted using only normal evaluation images. For category $c$, support seed $r$, and corruption $q$, each corrupted view is paired with its corresponding clean base image. The residual displacement is defined as
\[
\Delta s_{c,r,q}
=
\operatorname{median}_{x}
\log
\frac{s_{c,r}(x^{q})+\epsilon}
     {s_{c,r}(x^{\mathrm{clean}})+\epsilon},
\]
where $\epsilon=10^{-12}$ denotes a fixed numerical stabilizer. Since the paired views share identical support-derived calibration scores, this quantity measures corruption-induced movement relative to a fixed calibration reference and avoids comparisons of absolute residual scales across categories. The corresponding FAR inflation is
\[
\Delta\mathrm{FAR}_{c,r,q}
=
\mathrm{FAR}_{c,r,q}
-
\mathrm{FAR}_{c,r,\mathrm{clean}}.
\]
Both quantities are averaged over the five support seeds prior to computing Spearman's rank correlation across categories, separately for each dataset, $k$, and corruption. Categories, rather than images or seeds, serve as the analysis units. The correlations are descriptive and are not interpreted as causal relationships.

\subsection{Reproducibility and Claim Rules}
All experiments are conducted on a single NVIDIA RTX 5090 graphics processing unit (GPU) using PyTorch 2.12.1. DINOv2 patch features are cached in 32-bit floating-point (fp32) format. Reproducibility records include the declared protocol revision, recorded worktree state, DINOv2 source-tree and weight hashes, environment specifications, dataset counts, per-image predictions, base-image identities, support manifests, corruption parameters, source partitions, candidate-threshold bounds, and Secure Hash Algorithm 256-bit (SHA-256) checksums. Automated integrity checks are performed for 1,500 MVTec, 1,200 VisA, and 600 MPDD view cells, rejecting duplicate cells, missing support statistics, or overlap between support and test sets. Reliability methods cannot claim ranking improvement unless the raw score is altered; image-level and category-level certificates are not interchangeable.
\section{Results}\label{sec:results}

\subsection{Shift-Induced False-Alarm Inflation}\label{sec:target-only-operation}
\begin{table*}[pos=t]
\centering
\caption{Target-only LOIO operating points. The floor represents the smallest rank value, $1/(k{+}1)$, while the term 'attainable' refers to the largest rank-value grid point not exceeding the nominal $\alpha$, or zero if no such point exists. For each corruption, FAR and detection power are pooled across categories and seeds. The table subsequently averages the four corruption-specific rates.}
\label{tab:attainable-alpha}
\footnotesize
\setlength{\tabcolsep}{3pt}
\begin{tabular}{llrrrrr}
\toprule
dataset & $k$ & floor & nominal $\alpha$ & attainable & FAR & power \\
\midrule
\multirow{8}{*}{VisA (full)} & \multirow{4}{*}{4} & \multirow{4}{*}{0.200} & 0.01 & 0.000 & 0.000 & 0.000 \\
 & & & 0.05 & 0.000 & 0.000 & 0.000 \\
 & & & 0.10 & 0.000 & 0.000 & 0.000 \\
 & & & 0.20 & 0.200 & 0.142 & 0.595 \\
\cmidrule(lr){2-7}
 & \multirow{4}{*}{8} & \multirow{4}{*}{0.111} & 0.01 & 0.000 & 0.000 & 0.000 \\
 & & & 0.05 & 0.000 & 0.000 & 0.000 \\
 & & & 0.10 & 0.000 & 0.000 & 0.000 \\
 & & & 0.20 & 0.111 & 0.094 & 0.555 \\
\midrule
\multirow{8}{*}{MVTec (full)} & \multirow{4}{*}{4} & \multirow{4}{*}{0.200} & 0.01 & 0.000 & 0.000 & 0.000 \\
 & & & 0.05 & 0.000 & 0.000 & 0.000 \\
 & & & 0.10 & 0.000 & 0.000 & 0.000 \\
 & & & 0.20 & 0.200 & 0.278 & 0.880 \\
\cmidrule(lr){2-7}
 & \multirow{4}{*}{8} & \multirow{4}{*}{0.111} & 0.01 & 0.000 & 0.000 & 0.000 \\
 & & & 0.05 & 0.000 & 0.000 & 0.000 \\
 & & & 0.10 & 0.000 & 0.000 & 0.000 \\
 & & & 0.20 & 0.111 & 0.200 & 0.868 \\
\bottomrule
\end{tabular}
\end{table*}
\begin{table*}[pos=t]
\centering
\caption{Target-only LOIO FAR, detection power, and alarm precision at the first reported attainable level, $\alpha{=}0.20$. The results utilize five support-sampling seeds with support sets nested across $k$, and are pooled over images within each dataset, corruption, and $k$. All normal and anomalous test images are retained. Alarms are triggered when the rank value satisfies $p\le\alpha$, following the comparison tolerance specified in the text.}
\label{tab:false-alarm-control}
\footnotesize
\setlength{\tabcolsep}{3pt}
\begin{tabular}{llrrrr}
\toprule
dataset & corruption & $k$ & FAR & power & precision \\
\midrule
\multirow{8}{*}{VisA (full)} & \multirow{2}{*}{blur} & 4 & 0.144 & 0.606 & 0.825 \\
 & & 8 & 0.094 & 0.566 & 0.870 \\
\cmidrule(lr){2-6}
 & \multirow{2}{*}{bright/contr.} & 4 & 0.146 & 0.596 & 0.821 \\
 & & 8 & 0.098 & 0.557 & 0.864 \\
\cmidrule(lr){2-6}
 & \multirow{2}{*}{Gaussian noise} & 4 & 0.133 & 0.575 & 0.829 \\
 & & 8 & 0.081 & 0.538 & 0.881 \\
\cmidrule(lr){2-6}
 & \multirow{2}{*}{JPEG} & 4 & 0.145 & 0.602 & 0.822 \\
 & & 8 & 0.104 & 0.559 & 0.858 \\
\midrule
\multirow{8}{*}{MVTec (full)} & \multirow{2}{*}{blur} & 4 & 0.219 & 0.880 & 0.904 \\
 & & 8 & 0.139 & 0.870 & 0.936 \\
\cmidrule(lr){2-6}
 & \multirow{2}{*}{bright/contr.} & 4 & 0.214 & 0.882 & 0.906 \\
 & & 8 & 0.133 & 0.870 & 0.939 \\
\cmidrule(lr){2-6}
 & \multirow{2}{*}{Gaussian noise} & 4 & 0.341 & 0.873 & 0.857 \\
 & & 8 & 0.281 & 0.860 & 0.878 \\
\cmidrule(lr){2-6}
 & \multirow{2}{*}{JPEG} & 4 & 0.336 & 0.883 & 0.860 \\
 & & 8 & 0.248 & 0.872 & 0.892 \\
\bottomrule
\end{tabular}
\end{table*}

Table~\ref{tab:attainable-alpha} distinguishes unattainable nominal levels from observed false-alarm behavior. For $k=4$, the smallest rank value is $1/(k+1)=0.2$, so nominal levels of 0.01, 0.05, and 0.10 cannot trigger an alarm. When $k=8$, the floor decreases to $1/9\approx0.111$, yet these three nominal levels remain below the threshold. The resulting zero FAR and power values are artifacts of finite resolution, rather than evidence of conservative false-alarm control.

At a nominal $\alpha=0.20$, the effective threshold is 0.20 for $k=4$, but only $1/9$ for $k=8$, since $1/9$ is the largest $k=8$ grid point not exceeding 0.20. On VisA, the aggregate FAR and power pairs are 0.142/0.595 for $k=4$ and 0.094/0.555 for $k=8$. On MVTec, these values are 0.278/0.880 and 0.200/0.868, respectively. Aggregates are computed by first pooling images over categories and five support-sampling seeds within each corruption, followed by averaging the four corruption-specific rates. The MVTec $k=8$ average matches the requested 0.20 budget but exceeds the 0.111 tie-free discrete reference at its effective threshold. This does not constitute an aggregate violation of the requested budget, although the anticipated finite-grid conservatism is not observed. The nominal level produces different effective thresholds across $k$, while the empirical FAR is not constrained to the rank-value grid.

Table~\ref{tab:false-alarm-control} presents the aggregate FAR by corruption type. The headline value of 0.341 represents the empirical FAR for Gaussian-corrupted MVTec at $k=4$. This value is neither the average across MVTec nor a population-level false-alarm probability. It is the largest dataset-level aggregate in the frozen grid and is approximately 1.7 times the nominal value of 0.20. Category-level FARs display considerable heterogeneity: after pooling within each category over five seeds, the median is 0.215, the interquartile range spans 0.169 to 0.481, and 9 out of 15 categories exceed 0.20. Therefore, the aggregate is neither consistent across categories nor attributable to a single category. JPEG compression also exhibits anti-conservative behavior at 0.336. At $k=8$, blur and brightness/contrast are conservative at 0.139 and 0.133, respectively, while Gaussian noise and JPEG compression remain above the nominal threshold at 0.281 and 0.248. Increasing $k$ reduces, but does not eliminate, the observed mismatch.

VisA demonstrates distinct behavior: the FAR remains between 0.133 and 0.146 at $k=4$ and between 0.081 and 0.104 at $k=8$. This indicates that the shift effect is dependent on both the dataset and corruption condition, rather than representing a universal inflation. Power remains nonzero at all reported operating points, while precision is considered descriptive due to its dependence on the benchmark anomaly prevalence. Because fp32 and double-precision representations of 0.2 differ slightly, exact mixed-precision comparisons may inadvertently exclude floor-level alarms. All reported thresholding procedures employ the fixed numerical tolerance specified in the protocol.

\begin{figure*}[pos=t]
    \centering
    \includegraphics[width=0.92\linewidth]{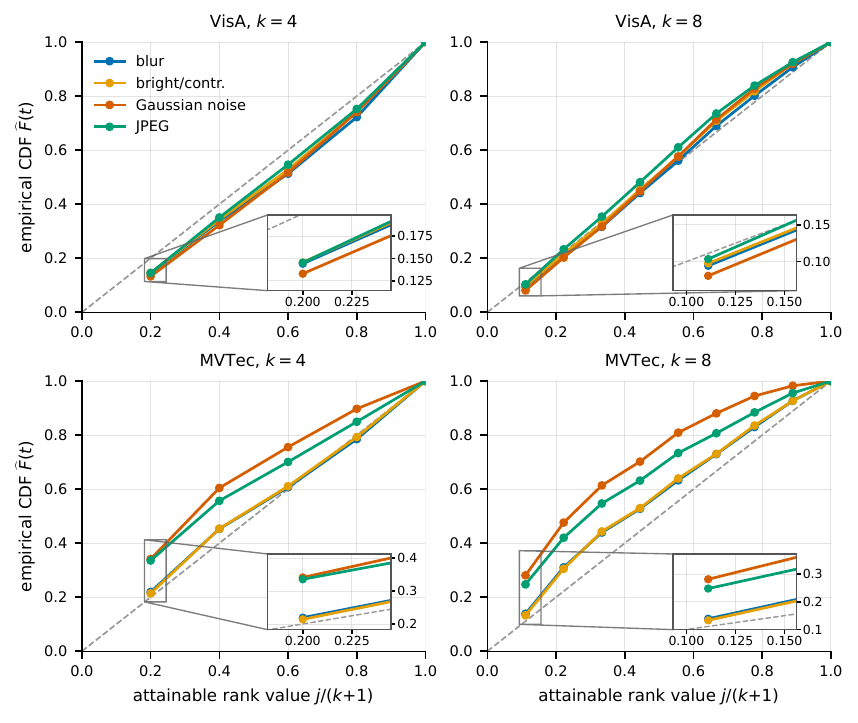}
    \caption{Empirical CDF of normal-image LOIO rank values at each attainable point $j/(k{+}1)$, compared to the tie-free discrete-uniform benchmark (dashed diagonal), using five support-sampling seeds. Valid ties may cause the population CDF to fall below this benchmark. Points above the diagonal indicate an excess of empirical false alarms relative to the benchmark. The lower-right insets and connector rays provide an enlarged view of the first attainable grid region.}
    \label{fig:uniformity}
\end{figure*}

The normal-rank audit clarifies the direction of these deviations. Under the relevant exchangeability conditions, a valid conformal p-value satisfies $P(p\leq\alpha)\leq\alpha$ for normal data. Equation~\eqref{eq:loio} does not assume these conditions; therefore, its conformal-form rank values are compared with this reference rather than presuming validity. Since a $k$-score rank value lies on the grid $\{j/(k+1)\}$, a continuous Kolmogorov–Smirnov test against $U(0,1)$ is not appropriate. Figure~\ref{fig:uniformity} instead compares the empirical normal-rank CDF with the tie-free discrete-uniform benchmark at each attainable point. Under valid exchangeability, ties can render the rank values more conservative, so the population CDF need only lie at or below this benchmark; the finite pooled CDF remains descriptive. For VisA at $k=4$, the results are conservative, with the largest observed negative gap, $-0.087$, occurring under blur at 0.6. At $k=8$, the first grid point remains conservative, although small anti-conservative gaps appear at higher grid points. MVTec is anti-conservative at both $k$ values, particularly under Gaussian noise and JPEG compression. For Gaussian-corrupted MVTec at $k=4$, $\widehat F(0.2)=0.341$ compared to the benchmark value of 0.20, and the gap increases to 0.204 at the 0.4 grid point.

Rows share calibration sets, categories, and repeated test images across seeds. Consequently, the CDF comparison is treated as descriptive, and an iid goodness-of-fit test is not applied to the pooled points. The results indicate that corrupted normal scores increase relative to clean support calibration scores, although no causal mechanism is identified. On the full MVTec dataset at $k=4$, the fold-matched LOIO sensitivity increases the clean FAR from 0.212 to 0.257 and power from 0.880 to 0.906. This trend aligns with the full-support test fit suppressing alarms relative to the $k-1$ calibration fits; the resulting asymmetry can make the original audit appear more conservative, but does not establish validity. Fold matching also fails to restore false-alarm control.

\subsection{Category-Level Residual Displacement}

The paired analysis establishes a relationship between dataset-level FAR contrast and the displacement of normal residual scores. In the MVTec dataset, both Gaussian noise and JPEG compression result in systematic positive shifts in scores. At $k=4$, the category-level Spearman correlations between residual displacement and FAR inflation are $\rho=0.825$ for Gaussian noise and $\rho=0.875$ for JPEG. At $k=8$, these correlations increase to $0.843$ and $0.914$, respectively. The corresponding median log-score displacements are 0.037 and 0.025 at $k=4$, and 0.042 and 0.027 at $k=8$. Median FAR inflation exhibits a similar trend: 0.033 and 0.036 at $k=4$, and 0.067 and 0.050 at $k=8$.

In contrast, the VisA dataset does not demonstrate a comparable dataset-wide displacement. For both Gaussian noise and JPEG compression, the median log-score shifts are approximately 0.000 and 0.004 at both support sizes. Median FAR changes are $-0.003$ and $-0.005$ at $k=4$, and $-0.007$ and 0.002 at $k=8$. Within VisA, the category-level correlations remain positive, suggesting that categories with greater score displacement tend to exhibit larger FAR changes. However, both metrics remain centered near zero at the dataset level.

These results support residual displacement as a proximal explanation for the observed FAR behavior. In MVTec, normal scores increase systematically under Gaussian noise and JPEG compression, whereas VisA normal scores exhibit minimal aggregate movement. The analysis does not address the underlying reasons for the differing responses of the frozen representation across datasets, nor does it establish a causal relationship with dataset composition. Since FAR is determined by thresholding ranks derived from the same residual scores, the observed correlation is partly structural. Its primary value lies in demonstrating that the aggregate contrast is distributed across categories and aligns with measurable score displacement, rather than being attributable to a single category.

\subsection{Category-Count Feasibility}\label{sec:category-feasibility-results}
Figure~\ref{fig:certificate-feasibility} and Table~\ref{tab:certificate-feasibility} present analytic design calculations rather than fitted learning curves. In the optimistic scenario without multiplicity, the all-zero distribution-free floor intersects $\alpha=0.20$, 0.10, and 0.05 at 14, 29, and 59 iid category draws. These counts are necessary under Proposition~3 but are not sufficient to achieve a positive CRESS threshold. Using the frozen family allocation with $A=3$ and favorable $M=1$, the lower limits increase to 22, 46, and 94, while the Hoeffding rule requires 60, 240, and 958 draws.

The presence of candidate multiplicity further increases the required sample sizes. With the cap at $M=20$, the family-adjusted distribution-free lower limits rise to 35, 74, and 152, while the Hoeffding requirements increase to 98, 390, and 1,557. The three or four certification categories available in the frozen audit remain below even the most optimistic 14-category requirement at $\alpha=0.20$. This limitation is not solely due to allocating 25\% of the source pool to certification. Any disjoint split with nonempty reference and proposal roles can allocate at most 12 MVTec, 9 VisA, or 13 transfer-source categories to certification, all below 14. This comparison addresses a feasibility question before evaluating target performance. Within the deterministic distribution-free framework of Proposition~3, a nonzero certificate of marginal new-category risk cannot be obtained from the current source pools, even when no category loss is observed. However, this result does not preclude parametric, hierarchical, randomized, or side-information-based procedures, which require separate assumptions and validity justifications.

The confidence-level perspective further illustrates the insufficiency. At $\alpha=0.20$, three all-zero categories provide a maximum of 48.8\% confidence, while four categories provide up to 59.0\% confidence under the optimistic multiplicity-free assumptions of Proposition~3. If a 95\% confidence level is required, the corresponding all-zero UCB floors are 0.632 and 0.527, respectively. Both values significantly exceed 0.20, even before accounting for candidate multiplicity or image-unit co-reporting.

\subsection{CRESS Boundary Test: Categories versus Images}\label{sec:cress-strict-results}
\begin{table*}[t]
\centering
\caption{Frozen strict CRESS audit over 960 job $\times$ source-view $\times$ $k$ $\times$ $\alpha$ $\times$ condition configurations, 240 per source-to-target job. Each configuration includes 75 MVTec, 60 VisA, or 30 MPDD target-category/seed cells; thus, the 'nonzero frac.' metric aggregates 18,000, 14,400, or 7,200 cells per respective job. 'Min UCB' denotes the smallest observed candidate bound. Category and image rows address distinct estimands; image rows operate under the idealized iid source-image model.}
\label{tab:strict-cress}
\footnotesize
\setlength{\tabcolsep}{5pt}
\begin{tabular}{llrrrl}
\toprule
source $\rightarrow$ target & certification unit & units/cell & min UCB & nonzero frac. & interpretation \\
\midrule
MVTec $\rightarrow$ MVTec & category & 4 & 0.950 & 0.000 & category-unit gate fails \\
VisA $\rightarrow$ VisA & category & 3 & 1.000 & 0.000 & category-unit gate fails \\
MVTec $\rightarrow$ VisA & category & 4 & 0.961 & 0.000 & category-unit gate fails \\
MVTec $\rightarrow$ MPDD & category & 4 & 0.986 & 0.000 & category-unit gate fails \\
\midrule
MVTec $\rightarrow$ MVTec & source image & 75 to 191 & 0.039 & 0.367 & iid-image sensitivity \\
VisA $\rightarrow$ VisA & source image & 150 to 180 & 0.042 & 0.603 & iid-image sensitivity \\
MVTec $\rightarrow$ VisA & source image & 72 to 159 & 0.048 & 0.415 & iid-image sensitivity \\
MVTec $\rightarrow$ MPDD & source image & 81 to 190 & 0.040 & 0.370 & iid-image sensitivity \\
\bottomrule
\end{tabular}
\end{table*}
\begin{table}[pos=t]
\centering
\caption{Direct pooled-source conformal under matched-condition selection. Reported entries represent unweighted means of within-target-cell rates across categories, $k\in\{1,2,4,8\}$, five support seeds, and five conditions. The complete selected source pool applies $\tau=\alpha$ without a reference/proposal/certification split; this is an empirical, uncertified baseline.}
\label{tab:pooled-source-conformal}
\footnotesize
\setlength{\tabcolsep}{4pt}
\begin{tabular}{lrrr}
\toprule
source $\rightarrow$ target & $\alpha$ & FAR & power \\
\midrule
\multirow{3}{*}{MVTec $\rightarrow$ MVTec}
 & 0.05 & 0.075 & 0.404 \\
 & 0.10 & 0.131 & 0.566 \\
 & 0.20 & 0.242 & 0.754 \\
\cmidrule(lr){1-4}
\multirow{3}{*}{VisA $\rightarrow$ VisA}
 & 0.05 & 0.060 & 0.320 \\
 & 0.10 & 0.111 & 0.466 \\
 & 0.20 & 0.208 & 0.639 \\
\cmidrule(lr){1-4}
\multirow{3}{*}{MVTec $\rightarrow$ VisA}
 & 0.05 & 0.028 & 0.171 \\
 & 0.10 & 0.048 & 0.282 \\
 & 0.20 & 0.099 & 0.441 \\
\cmidrule(lr){1-4}
\multirow{3}{*}{MVTec $\rightarrow$ MPDD}
 & 0.05 & 0.030 & 0.186 \\
 & 0.10 & 0.066 & 0.283 \\
 & 0.20 & 0.157 & 0.423 \\
\bottomrule
\end{tabular}
\end{table}

Table~\ref{tab:strict-cress} evaluates whether the analytic shortage constraint is active in the frozen CRESS protocol. The 960 gate configurations consist of four source-to-target jobs, four source-view modes, four $k$ values, three $\alpha$ levels, and five conditions, resulting in 240 configurations per job; each aggregates target categories and five support seeds. These are repeated configurations within a fixed grid, not independent statistical trials, and Proposition~1 provides a per-cell rather than a grid-wide simultaneous guarantee. Category-unit certification assigns $\tau^\star=0$ to every underlying cell, and no configuration achieves the required 80\% nonzero-threshold rate. The smallest candidate UCB is 0.950 for MVTec, 1.000 for VisA, 0.961 for MVTec-to-VisA, and 0.986 for transfer from MVTec to MPDD. Even the smallest value is 4.75 times greater than the largest tested $\alpha=0.20$.

The fail-closed outcome aligns with the design calculation and does not indicate a computational or optimization error. Proposition~2 stipulates a minimum of 60 categories for the favorable $M=1$ Hoeffding gate at $\alpha=0.20$, while Proposition~3 requires at least 14 categories in the optimistic multiplicity-free, distribution-free scenario and at least 22 under the favorable frozen allocation. Only three or four certification categories are available. Since the operational gate is conjunctive, the zero nonzero-threshold rate directly determines failure; subsequent target-performance checks cannot transform $\tau^\star=0$ into a usable alarm threshold. This zero-threshold result persists even when $k=1$ stress cells are excluded and only $k\in\{2,4,8\}$ is considered, indicating independence from the patch-split fallback. The strict experiment thus demonstrates a fail-closed outcome consistent with the predicted category-count boundary across the entire grid, rather than providing 960 independent replications of the theoretical result.

The image-unit analysis alters both the sample size and the estimand. Under the idealized conditional iid source-image assumption, 72 to 191 selected source-image units per cell reduce the smallest UCB to a range of 0.039 to 0.048. Positive thresholds are observed in 36.7\% of MVTec, 60.3\% of VisA, 41.5\% of MVTec-to-VisA, and 37.0\% of MVTec-to-MPDD target cells. These rates remain below the 80\% operational gate threshold. Importantly, the associated risk pertains to image draws from the selected source mixture. The inclusion of additional images does not generate independent categories and cannot be interpreted as evidence for a marginal new-category certificate. Consequently, the same source data may be informative for the image-mixture estimand but insufficient for a distribution-free category-transfer estimand.

Table~\ref{tab:pooled-source-conformal} presents direct pooled-source conformal analysis as an uncertified empirical reference. This approach utilizes the complete selected source pool with $\tau=\alpha$ and does not partition data into reference, proposal, or certification categories. Reported entries are unweighted means of within-target-cell rates, ensuring equal weighting for categories and support seeds regardless of image count. Under matched-condition selection, the mean FAR exceeds the nominal value within MVTec and VisA at all displayed levels; transfer scenarios are more conservative and less powerful. Direct pooling provides finer operating points than target-only LOIO, but characterizes empirical risk under the selected source mixture rather than certified new-category risk.

\subsection{Supporting Ranking Substrate and Storage}
\begin{table*}[pos=t]
\centering
\caption{Clean image-level ranking performance and category-specific ranker state in mebibytes (MiB). The controlled DINOv2 NN and PCA64/PCA128 methods utilize a shared feature set, data split, evaluation metric, and storage-accounting protocol, with results averaged over categories and five random seeds. AnomalyDINO-S (448) and WinCLIP values are reported in~\cite{damm2025anomalydino,winclip2023} under their respective protocols. Bold marks the best value only among common-protocol rows within each dataset/$k$ block. The notation '--' indicates not evaluated, N/C denotes not directly comparable, and N/R signifies not reported.}
\label{tab:clean-efficiency}
\setlength{\tabcolsep}{2.4pt}
\footnotesize
\begin{tabular}{lllrrr}
\toprule
Dataset & $k$ & Method & AUROC $\uparrow$ & AP $\uparrow$ & MiB $\downarrow$ \\
\midrule
\multirow{11}{*}{MVTec}
& \multirow{4}{*}{1} & AnomalyDINO-S (reported) & 0.965 & 0.981 & -- \\
& & WinCLIP (reported) & 0.931 & N/C & N/R \\
& & Controlled DINOv2 NN & \textbf{0.914} & \textbf{0.952} & 2.005 \\
& & PCA64 (controlled) & 0.904 & 0.951 & \textbf{0.095} \\
\cmidrule(lr){2-6}
& \multirow{4}{*}{4} & AnomalyDINO-S (reported) & 0.976 & 0.984 & -- \\
& & WinCLIP (reported) & 0.952 & N/C & N/R \\
& & Controlled DINOv2 NN & \textbf{0.942} & \textbf{0.967} & 6.000 \\
& & PCA64 (controlled) & 0.937 & \textbf{0.967} & \textbf{0.095} \\
\cmidrule(lr){2-6}
& \multirow{3}{*}{8} & AnomalyDINO-S (reported) & 0.980 & 0.990 & -- \\
& & Controlled DINOv2 NN & \textbf{0.948} & 0.970 & 6.000 \\
& & PCA64 (controlled) & 0.945 & \textbf{0.972} & \textbf{0.095} \\
\midrule
\multirow{14}{*}{VisA}
& \multirow{5}{*}{1} & AnomalyDINO-S (reported) & 0.856 & 0.866 & -- \\
& & WinCLIP (reported) & 0.838 & N/C & N/R \\
& & Controlled DINOv2 NN & 0.804 & 0.809 & 2.005 \\
& & PCA64 (controlled) & 0.823 & 0.834 & \textbf{0.095} \\
& & PCA128 (controlled) & \textbf{0.834} & \textbf{0.842} & 0.189 \\
\cmidrule(lr){2-6}
& \multirow{5}{*}{4} & AnomalyDINO-S (reported) & 0.913 & 0.918 & -- \\
& & WinCLIP (reported) & 0.873 & N/C & N/R \\
& & Controlled DINOv2 NN & 0.862 & 0.861 & 6.000 \\
& & PCA64 (controlled) & 0.870 & 0.883 & \textbf{0.095} \\
& & PCA128 (controlled) & \textbf{0.885} & \textbf{0.895} & 0.189 \\
\cmidrule(lr){2-6}
& \multirow{4}{*}{8} & AnomalyDINO-S (reported) & 0.926 & 0.929 & -- \\
& & Controlled DINOv2 NN & 0.873 & 0.870 & 6.000 \\
& & PCA64 (controlled) & 0.882 & 0.894 & \textbf{0.095} \\
& & PCA128 (controlled) & \textbf{0.897} & \textbf{0.905} & 0.189 \\
\bottomrule
\end{tabular}
\end{table*}

Table~\ref{tab:clean-efficiency} demonstrates that the reliability conclusions are based on a viable and compact ranking substrate. The AnomalyDINO-S (448) results are reported using the method's published protocol and serve as external accuracy references, rather than as evidence of superiority under the controlled protocol. Similarly, WinCLIP is referenced for its reported AUROC and area under the precision-recall curve (AUPR) at one and four shots; however, it does not provide an AP implementation verified to match the current protocol. Accordingly, its AP entries are marked as N/C to avoid conflating reported AUPR values with the controlled AP comparison. Additionally, WinCLIP does not report category-specific storage under the current accounting method or provide an eight-shot result.

Compared to the controlled DINOv2 NN ranker, PCA64 achieves lower AUROC by 0.011, 0.005, and 0.003 on MVTec, and higher AUROC by 0.019, 0.008, and 0.009 on VisA at $k=1,4,8$, respectively. PCA64 serves as a compact supporting ranker and does not represent a novel ranking contribution or a state-of-the-art claim. On VisA, PCA128, evaluated exclusively, improves mean AUROC over PCA64 by 0.011, 0.016, and 0.014. All differences are calculated using unrounded values. Boldface is used to compare only common-protocol rows, excluding external references.

\emph{Storage accounting.} Storage calculations exclude the shared frozen DINOv2 ViT-S/14 backbone. For $d=384$ fp32 features, the PCA mean and basis require $(m+1)d$ scalars, corresponding to 0.095\,MiB for PCA64 and 0.189\,MiB for PCA128. Historical PCA128 artifacts include a 0.566\,MiB calibration wrapper; however, raw ranking utilizes only PCA residuals, so Table~\ref{tab:clean-efficiency} reports the 0.189\,MiB ranker state. Support-normalization scalars are negligible at the reported precision. The controlled memory bank stores 2.005\,MiB at $k=1$ and is capped at 4096 vectors, or 6.000\,MiB, for higher values. CRESS retains normalized scalar reference scores and partition metadata instead of source patch features. No runtime claims are made.

The clean benchmark employs max pooling, while the conformal pipeline utilizes the top-$1\%$ mean as a smoother operational score. Table~\ref{tab:agg-ablation} examines this choice across three MVTec and four VisA categories over three seeds. Among the displayed rows, the largest AUROC difference between either paper's choice and the best tested aggregator is 0.018. Moderate top-fraction pooling outperforms max pooling on MVTec, while top-$0.5\%$ is optimal on VisA. This limited subset supports the adoption of a smoothed operational score but does not establish invariance to aggregation across the full dataset. Consequently, Table~\ref{tab:clean-efficiency} demonstrates that the inherited substrate is compact and viable under the specified clean protocol, but it is not used to attribute any reliability difference to the PCA ranker.

\begin{table}[t]
\centering
\caption{Aggregation-sensitivity ablation for clean image AUROC under max and top-$\rho$ mean pooling. Results average three MVTec categories (bottle, cable, and hazelnut) or four VisA categories (candle, cashew, pcb1, and pipe fryum) over seeds from 0 to 2. Bold marks the best tested aggregator in each row. This subset analysis diagnoses sensitivity within the displayed grid; it is not a full-dataset ranking comparison.}
\label{tab:agg-ablation}
\footnotesize
\setlength{\tabcolsep}{4pt}
\begin{tabular}{llrrrrr}
\toprule
dataset & $k$ & max & top-0.5\% & top-1\% & top-2\% & top-5\% \\
\midrule
\multirow{2}{*}{MVTec} & 4 & 0.9571 & 0.9638 & 0.9667 & \textbf{0.9693} & 0.9621 \\
 & 8 & 0.9578 & 0.9685 & 0.9740 & \textbf{0.9759} & 0.9727 \\
\midrule
\multirow{2}{*}{VisA} & 4 & 0.9039 & \textbf{0.9101} & 0.9038 & 0.8944 & 0.8806 \\
 & 8 & 0.9101 & \textbf{0.9200} & 0.9147 & 0.9071 & 0.8941 \\
\bottomrule
\end{tabular}
\end{table}
\FloatBarrier

\section{Limitations}\label{sec:limitations}
The guarantees apply within a defined statistical scope. Proposition~3 addresses deterministic, uniformly valid, distribution free UCBs based only on iid bounded category losses. The counts 14, 29, and 59 represent optimistic necessary limits without accounting for multiplicity, and do not constitute sufficient CRESS budgets. Methods utilizing pre-specified parametric or hierarchical models, informative auxiliary data, or randomized procedures may function with fewer categories, but each requires a distinct validity analysis. Proposition~1 governs the declared candidate family within a single fixed experimental cell and does not claim a single simultaneous 95\% event across the entire reported grid.

The empirical conclusions are directly linked to the specified sampling unit and benchmark scope. At the category level, interpretation considers entire category units, including support construction and held-out views, as iid samples from a defined meta-population. In contrast, image-unit sensitivity focuses on the selected mixture of source images. Successful transfer to a realized target further requires source-risk dominance, whereas sampling from the same meta-population supports only marginal statements about new categories. The audit encompasses MVTec, VisA, MPDD, and four controlled corruptions, rather than the full spectrum of production shifts. The frozen DINOv2 PCA residual serves as an inherited ranking substrate, not as a new state-of-the-art ranker. These limitations underscore the need for validation on category-rich source archives and anticipated operational shifts, as well as for assumption-aware alternatives to the distribution-free bounds examined in this study.

\section{Conclusion and Future Work}\label{sec:conclusion}
This study has established that the reliability of few-shot anomaly thresholds depends on both the amount and the statistical unit of normal evidence. At nominal $\alpha=0.20$, the target-only LOIO audit has yielded an empirical FAR of 0.341 on Gaussian-corrupted MVTec at $k=4$, showing that an attainable operating point need not remain stable under shift. The finite floor $1/(k+1)$ has explained when a target-only rule is structurally silent, while the shift audit has shown that crossing this floor does not supply an exchangeability guarantee.

The category-count analysis has quantified the separate obstacle faced by source-assisted certification. It has shown that, even with all category losses zero, optimistic multiplicity-free certification by a deterministic, uniformly valid, distribution-free 95\% UCB requires at least 14, 29, and 59 iid category draws at $\alpha=0.20$, 0.10, and 0.05. The strict audit has provided only three or four certification categories, and even a certification-heavy reallocation with nonempty reference and proposal roles could provide at most 9 to 13. Category-unit CRESS has therefore returned $\tau^\star=0$ throughout the 960 frozen configurations, consistent with the feasibility boundary rather than 960 independent statistical failures. By contrast, image-unit sensitivity analyses have selected positive thresholds in 36.7\% to 60.3\% of target cells under the idealized selected-mixture iid model, confirming that images and categories purchase evidence for different estimands.

The resulting contribution is a practical study-design discipline: distinguish ranking from threshold reliability, identify attainable operating points, define the certification unit before evaluation, and match each guarantee to its estimand. Future work should evaluate category-rich source archives, real production shifts, temporal dependence, and condition-estimation error, and should investigate parametric, hierarchical, or side-information models whose additional assumptions are independently validated. Such work can convert the feasibility calculus developed here into more informative certification procedures without obscuring the distinction between image-mixture, marginal new-category, and category-conditional risk.
\section*{Declaration of competing interest}
The authors declare that they have no known competing financial interests or personal relationships that could have appeared to influence the work reported in this paper.

\section*{Data availability}
The datasets analyzed during the current study are publicly available from the \href{https://www.mvtec.com/company/research/datasets/mvtec-ad}{MVTec AD dataset page}, the \href{https://github.com/amazon-science/spot-diff}{VisA repository}, and the \href{https://github.com/stepanje/MPDD}{MPDD repository}.

\section*{Code availability}
The experimental code, corruption pipeline, integrity audits, and table and figure generators are available in \href{https://github.com/OxyzGiaHuy/CRESS}{CRESS}.

\section*{Funding}
This research did not receive any specific grant from funding agencies in the public, commercial, or not-for-profit sectors.

\section*{Declaration of generative AI and AI-assisted technologies in the manuscript preparation process}
During the preparation of this work, the authors used OpenAI Codex to support code review, software-test generation, reproducibility checks, statistical claim auditing, manuscript structuring and drafting, and language revision. The authors reviewed and edited all tool-assisted material and take full responsibility for the content of the publication.

\bibliographystyle{cas-model2-names}
\bibliography{references}

\end{document}